\documentclass[journal]{IEEEtran}
\ifCLASSINFOpdf
   \usepackage[pdftex]{graphicx}
   \graphicspath{{../pdf/}{../jpeg/}}
   \DeclareGraphicsExtensions{.pdf,.jpeg,.png}
\else
\fi
\usepackage{amsmath}
\usepackage{booktabs}

\begin{document}
%
\title{Effective Medical Code Prediction \\via Label Internal Alignment}
%
%
%

\author{Guodong~Liu
\thanks{Department of Electrical and Computer Engineering, University of pittsburgh USA e-mail: (guodong.liu.e@pitt.edu).}
}

%
%

\markboth{Journal of \LaTeX\ Class Files,~Vol.~14, No.~8, August~2015}%
{Shell \MakeLowercase{\textit{et al.}}: Bare Demo of IEEEtran.cls for IEEE Journals}
%



\maketitle

\begin{abstract}
The clinical notes are usually typed into the system by physicians. They are typically required to be marked by standard medical codes, and each code represents a diagnosis or medical treatment procedure. Annotating these notes is time consuming and prone to error. In this paper, we proposed a multi-view attention based Neural network to predict medical codes from clinical texts. Our method incorporates three aspects of information, the semantic context of the clinical text, the relationship among the label (medical codes) space, and the alignment between each pair of a clinical text and medical code. Our method is verified to be effective on the open source dataset. The experimental result shows that our method achieves better performance against the prior state-of-art on multiple metrics.
\end{abstract}


%
\IEEEpeerreviewmaketitle

\section{Introduction}
%
%
%
%
\IEEEPARstart{T}{he} electronic health records (EHR) system is widely accepted and deployed. Typically, certain collections of International Classification of Diseases (ICD) are encoded into patient's health reports. The ICDs are a set of standard alphanumeric codes used by physicians, data scientists\cite{choi2016doctor}, health insurance companies, and public health agencies across the world to indicate diagnoses and procedures\cite{o2005measuring}.

Traditionally, an ICD code is annotated by the coder of the hospital's Medical Record Department according to a doctor's clinical diagnosis record. To implement this task, the coder is required to be trained in coding rules, medicine, medical terminologies or other prior knowledge, which makes manually coding to be an expensive, time-consuming and error-prone work.  Thus, automatic coding will  benefit a large range of industries in various aspects. Apart from reducing the labor of physicians in recording and statistician in analysis, it would also facilitate the communication with other health related fields such as processing health insurance claims, tracking disease epidemics and compiling worldwide mortality statistics\footnote{https://www.verywellhealth.com/icd-10-codes-and-how-do-they-work-1738471}. As a truth, a growing number of professionals are seeking the automatically ICD coding mechanism since at least the 1990s\cite{de1998hierarchical}.

Generally speaking, the ICD prediction shows three characteristics. First, the label space is large. The ICD-9 contains over 15,000 codes, and this number is even over 140,000 in the ICD-10\cite{world2016icd}. Second, the data is unstructured. The clinical notes are stored in a free text format. Third, clinical texts involve inexact expressions, such as, misspelling and informal abbreviations. These features combined to bring a great challenge to machine learning algorithms and human coders alike.

As a multi-labels classification problem, the automatic ICD prediction has been explored in various classic machine learning branches, \emph{e.g.}, support vector machine~\cite{ferrao2013using}, $k$-nearest neighbors~\cite{ruch2008episodes,erraguntla2012inference}, Naive Bayes~\cite{pakhomov2006automating,medori2010machine}~\emph{etc}. Although some good results have been achieved, their applications are still limited. The main obstacle lies on the aforementioned unstructured free text input. Because the performance of these classical methods largely relies on the hand-crafted features (feature engineering), whereas, it is extremely difficult to extract efficient features from the plain clinic notes,  even for the most experienced experts in this field. Consequently, the research on end-to-end feature engineering attracts more attentions in the machine learning community.

During the recent decade, neural models have become the mainstream of learning underlying representations of unstructured data, which benefit from accumulated data, upgraded hardware, and especially new proposed neural network models. Recurrent neural networks (RNNs)\cite{mikolov2010recurrent}including its variants~\emph{e.g.} LSTM\cite{hochreiter1997long}, make progress on learning the representation of clinical notes\cite{lipton2015phenotyping,choi2016doctor}. Unfortunately, RNNs involve recursive structures, they cannot be very efficiently parallelized yielding significant computational cost. Convolutional neural networks (CNNs) naturally enjoy the advantages of parallel computing, which can fully leverage the strength of Graphical Process Unit(GPU). Thereby, recent methods usually leverage CNN modules to learn the underlying representation of clinical notes \cite{mullenbach2018explainable,zhang2017deconvolutional}. To further enrich the representation of input text, attention mechanism\cite{bahdanau2014neural} and label embedding method are introduced as a standard component \cite{shi2017towards,wang2018joint}. Attention modules help to capture most informative snippets and make a great progress on ICD prediction tasks.\emph{e.g.} Shi~\emph{et al.}~\cite{shi2017towards} propose a hierarchical LSTM attention architecture which can automatically assign ICD diagnostic codes. Label embedding combined with attention works as a soft alignment between the word and the label representation. Wang~\emph{et al.}\cite{wang2018joint}propose a label-embedding attentive model for text classification. It embeds the words and labels with the same joint space, and measures the compatibility of word-label pairs to attend the document representations. Especially, Mullenbach~\emph{et al.}~\cite{mullenbach2018explainable} achieves the state-of-art performance on MIMIC-III open source database\cite{johnson2016mimic} by using an attention to link the document CNN based representation space and the label embedding space.

We notice that most of the state-of-art works focus on getting a better representation via alignments among word-label pairs\cite{wang2018joint}, or seeking a better label embedding initialization by encoding ICD code descriptions \cite{wang2018joint,mullenbach2018explainable,shi2017towards}. 
However, there is little attention paid on the labels' internal correlation. We try to fill this gap by proposing a Multi-view Alignment Model (MVAM) in this paper. We integrate a novel self-attention encoder module on the label embedding end. MVAM initializes label embedding randomly, and deploys the self-attention\cite{vaswani2017attention} module to construct a context related representation of the labels (medical codes). Our model is intrigued by the intuition that the label space can be independent from the wording embedding. In addition, there should be some more direct semantic structures among different labels, rather than indirect related through document word embedding. In order to deal with multi-labels case, we also use a per-label attention between the document and label to force the model to learn label-wise document representations. Considering aforementioned difficulties(large amount of ICD codes), we apply positional embedding to our label encoder module, which can help to lean better label embedding in high dimensional label space. We evaluate our method on two versions of  MIMIC-III\cite{johnson2016mimic}, a widely used open dataset of ICU medical records. Our model outperforms the previous state-of-art results from both full label and high frequency settings.

Our multi-view alignment model provides an elegant solution to the problems mentioned in clinical code prediction scenarios. The MVAM framework has the following salutary properties:
\begin{itemize}
\item \textbf{Scalability}. Our label self attention module can help to extend the good performance of MVAM to higher dimensional setting.
\item \textbf{Reliability}. MVAM runs in an end-to-end mode, it accepts free text input and makes the clinical codes prediction automatically. Its performance will not rely on the hand-craft feature anymore.
\item \textbf{Robustness}. The attention between the document words and each label captures the informative snippets correlated with a code's presence. This mechanism makes a contribution to counteracting the noise introduced by inexact expressions.
\end{itemize}
In addition to solving the specific problems, we have the following engineering considerations:
\begin{itemize}
\item \textbf{Flexibility}: The design of pipe line decouples the document encoder and the label encoder, they can be implemented in totally different structures.
\item \textbf{Effectiveness}. This novel model achieves the state-of-art results on a widely used open dataset over various metrics.
\item \textbf{Efficiency}: This framework is well designed to avoid any recurrent structure, and apply transformer\cite{vaswani2017attention} alike architecture instead, which is efficient for parallel computing. Also,the MVAM abandons any weight decay regularization, which means it has less hyper parameters to tune. Therefore, it is efficient on the training process.
\end{itemize}


\section{Model}
The ICD code prediction is treated as a multi-label document classification problem \cite{tsoumakas2007multi}. We first define this task in a formal way. $\mathcal{L}$ denotes the whole ICD-9 code set, $\mathcal{X}$ denotes the input clinical dataset, $|\mathbf{Y}|$ denotes all medical codes included in the dataset, thus $\mathbf{Y}$ is a subset of $\mathcal{L}$. The problem is to determine a value of
$\hat{y}_{i,l}\in \{0,1\}$, when given the $i^{th}$ sample $\mathbf{X}\subset{\mathcal{X}}$ (note that $X$ is a document). 
\begin{figure}[!tb]
  \flushleft
  \includegraphics[width=.48\textwidth]{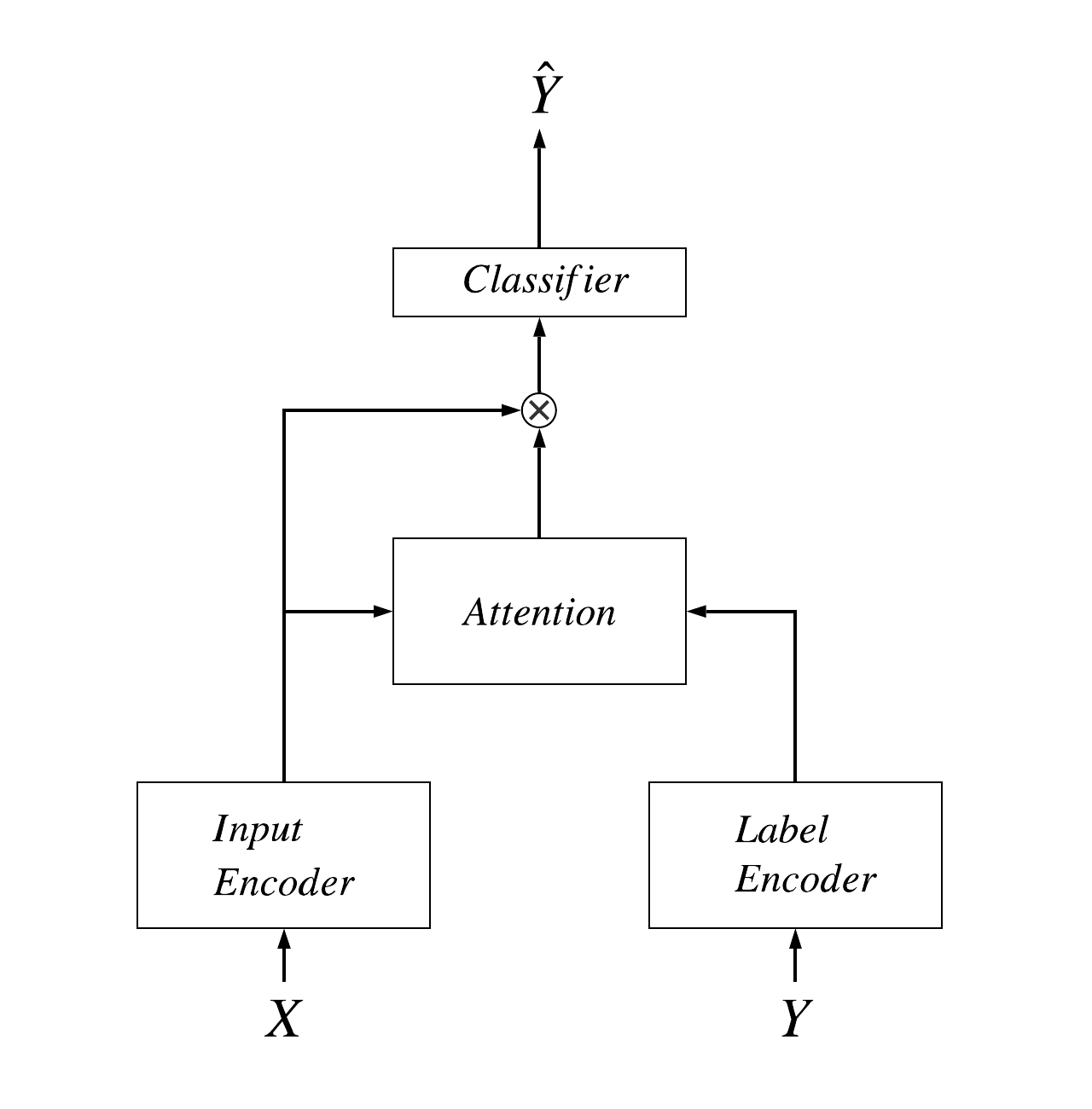}
  \caption{The pipe line of MVAM.}
  \label{figure1}
\end{figure}

We design an encoder-decoder structure pipe line as shown in figure \ref{figure1}. The input encoder accepts clinical text and encodes it into a basic representation. The label encoder accepts initial label embedding as input and calculates the representation via internal alignment. We use the output of encoders to compute alignment scores based on attention between the input and label representation. A final representation feeds into a classification layer ,the decoder in our framework, to make prediction. Note that we do not make any specific restriction on the implementation of any modules, and the only assumption is that a self-alignment module should be included in the label encoder. 

Taking into account parallel computing and simple principle , we decide to utilize CNN as input encoder and self-attention as label encoder. The attention module uses inner product to compute the comparability scores of words in $x_i$ under the $l^{th}$ label assumption. Rather than a common max pooling based strategy to distill the document representation, we use attention scores as weights to average along basic representation from input encoder. This policy is expected to select more informative semantic snippets in texts. Also compared with max pooling\cite{kim2014convolutional} or average pooling, it performs better on the long distance dependency. The weighted result will pass through the final classification layer which works as a decoder in a standard encoder-decoder structure. According to the simple principle\cite{shen2018baseline}, we give up all fancy decoders, and just employ $|\mathcal{L}|$ sigmoid classification units instead. We will introduce every detail of these modules as follows.

\subsection{Convolutional input encoder}
The text input sequence $\mathbf{X}$ goes through the embedding layer and generates a matrix $\mathbf{E}\in{\mathcal{R}^{{d_e}\times{N}}}=[e_1, e_2,...,e_N]$, where each column $e_i\in{\mathcal{R}^{d_e}}$ represents the pre-trained embedding for the original word $X_i$, and $N$ is the length of the input sequence $\mathbf{X}$, $d_e$ is the dimension of the pre-trained word embedding.

Upon the embedding layer, we leverage the Convolutional layer as the core module of clinical description encoder. A standard 1-dimensional CNN is applied to our practice. Denote the filter kernel as $\mathbf{W}_c\in{\mathcal{R}^{k\times{d_e}\times{d_c}}}$, where $k$ is the filter width, $d_e$ is the dimension of word embedding, and $d_c$ is the number of feature maps. Then at any position $i$, we have:
\begin{equation}
    h_i = \varphi ( \mathbf{W}_c * {e}_{i : i+k-1} + b_c )\,,
\end{equation}
where $*$ indicates the convolutional operator, $b_c\in \mathcal{R}^{d_c}$ is an bias, $\varphi$ denotes an element-wise nonlinear activation function.
In order to guarantee the output matrix $\mathbf{H}\in \mathcal{R}^{d_c \times N}$ has same sequence length $N$ with the original input, we pad $\mathbf{X}$ with zeros on each side.

\subsection{Self-attention ICD code encoder}
 There is an abundance of literature in the NLP community on label embedding\cite{mullenbach2018explainable,wang2018joint}, and most of them are focusing on either mapping the label space into the input word embedding space via various alignment strategies \emph{e.g.} attention, or mapping the input and label into the same representation space by homogeneous encoders\cite{shickel2017deep}. There is little research on designing any specific encoder to utilize interaction among labels. To fulfill this gap, we propose a transformer-alike\cite{vaswani2017attention} self-attention encoder to distill more informative representations to each medical code, and this is the main contribution of this paper. 
 
\begin{figure}[!tb]
  \centering
  \includegraphics[width=0.49\textwidth]{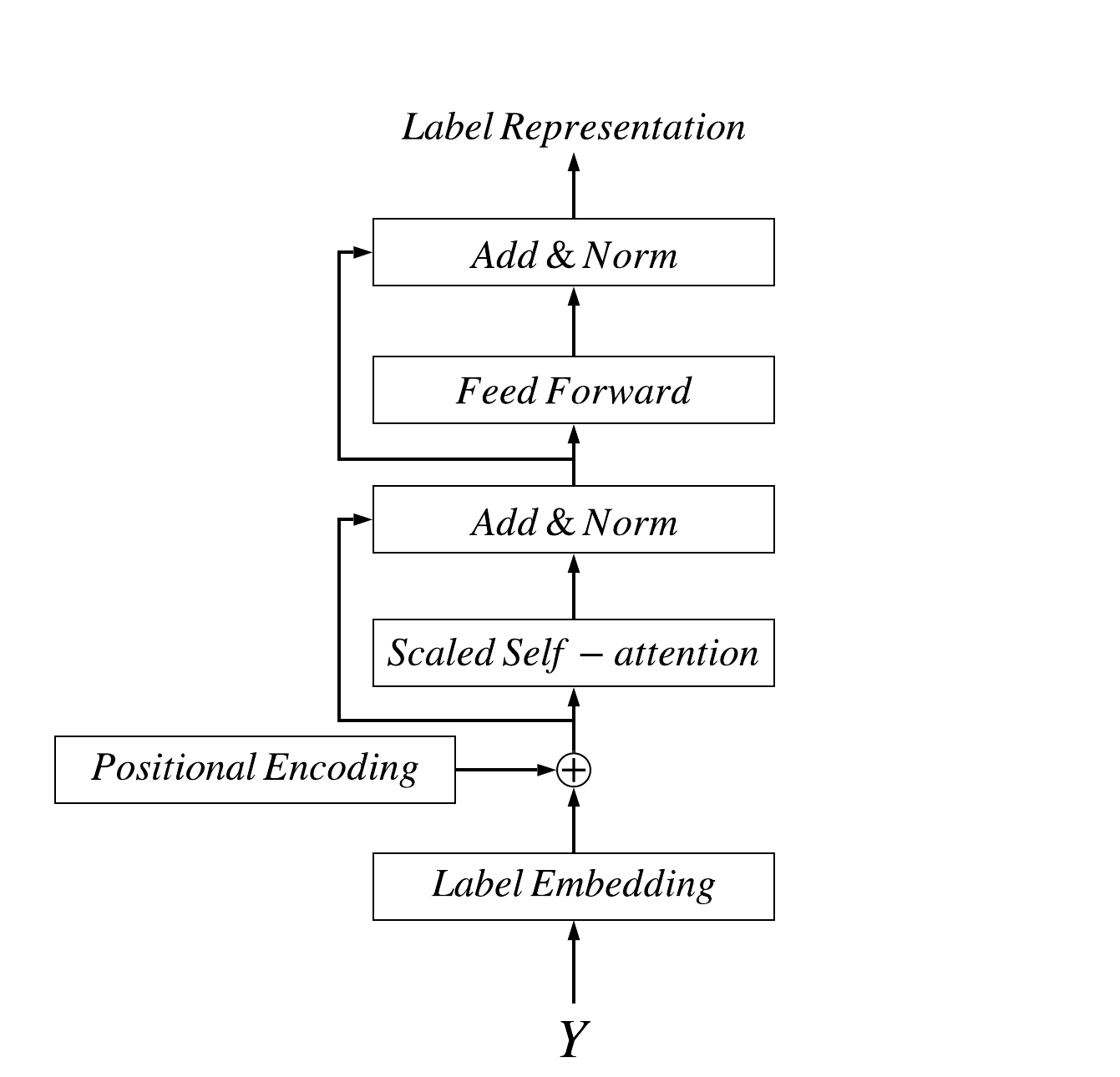}
  \caption{The architecture of self-attention label encoder.}
  \label{figure2}
\end{figure}

 The basic intuition is that clinical codes could include more semantic information than normal labels, \emph{e.g.} cat, dogs in images classification. Thus, they could convey internal relationship by themselves rather than fully relying on the manual crafted description of ICD codes. So we introduce a self-attention module to distill the internal interactions among all labels. The architecture of our self-attention label encoder is depicted in figure \ref{figure2}.
\subsubsection{\bfseries{Label embedding layer}}
$\mathbf{Y}$ denotes the original clinical codes \emph{e.g.} 292.0 is the ICD code of Drug withdrawal syndrome. The input is followed with a randomly initialized embedding layer, which is a simple trainable lookup table that stores embedding of a fixed dictionary $\mathbf{Y}$ and size $|Y|$. Note that this component can be replaced by any initialization.

This is also an advantage of our method. We do not require any other source of textual annotation for the clinical codes. On the contrary, other label embedding algorithms usually learn it via CNN or LSTM encoders. 

\subsubsection{\bfseries{Positional encoding}} Alike the aforementioned embedding layer, this module is also a trainable lookup table and has the same dimension with the label embedding layer. It stores the positional information of the encountered labels instead. The positional encoding values are added to the embedding matrix, providing information about structure of label space to MVAM. We initialize it by sine and cosine functions of different frequencies as depicted in\cite{vaswani2017attention}:
\begin{equation}
\begin{split}
    PE_{(pos, 2j)} = sin(pos/10000^{2j/d_e})\\
    PE_{(pos, 2j+1)} = cos(pos/10000^{2j/d_e})
\end{split}
\end{equation}
where $pos \in {(0,|\mathbf{Y}|)}$ is the position and $j\in{(0, \lfloor \frac{d_e}{2} \rfloor)}$ is the dimension. In \cite{vaswani2017attention}, the author claims that the this encoding would allow the model to easily learn to attend by relative positions, since for any fixed offset $k$, $PE_{pos+k}$ can be represented as a liner function of $PE_{pos}$.
We experimented both fixed and trainable positional encoding, they got no difference results, thus we decide to fix the positional encoding to keep our model efficient.

\subsubsection{\bfseries{Scaled dot-product self attention}} The sum of positional encoding and label embedding feeds into this module to learning a final representation of labels. We employ the scaled dot-product self attention\cite{vaswani2017attention} with the consideration of the scalability, effectiveness and efficiency. Compared with LSTM based encoder\cite{shickel2017deep}, self-attention's training does not rely on the previous states, which is easy to parallelize. Also the path length of long-range dependencies ($n$) in the network is $O(1)$ in our setting\cite{vaswani2017attention}, while it is $O(log_k(n))$ for a multi-layer CNN model with kernel width $k$. 

We can interpret the path length as how many layers that we need to build up a direct relation between any pairs of label with a distance $n$. Thus for a single layer CNN, it is hard to learn a direct correlation between pairs of labels with a distance larger that $k$. We will discuss the detail in the later sections. 

\begin{figure}[!tb]
  \centering
  \includegraphics[width=0.4\textwidth]{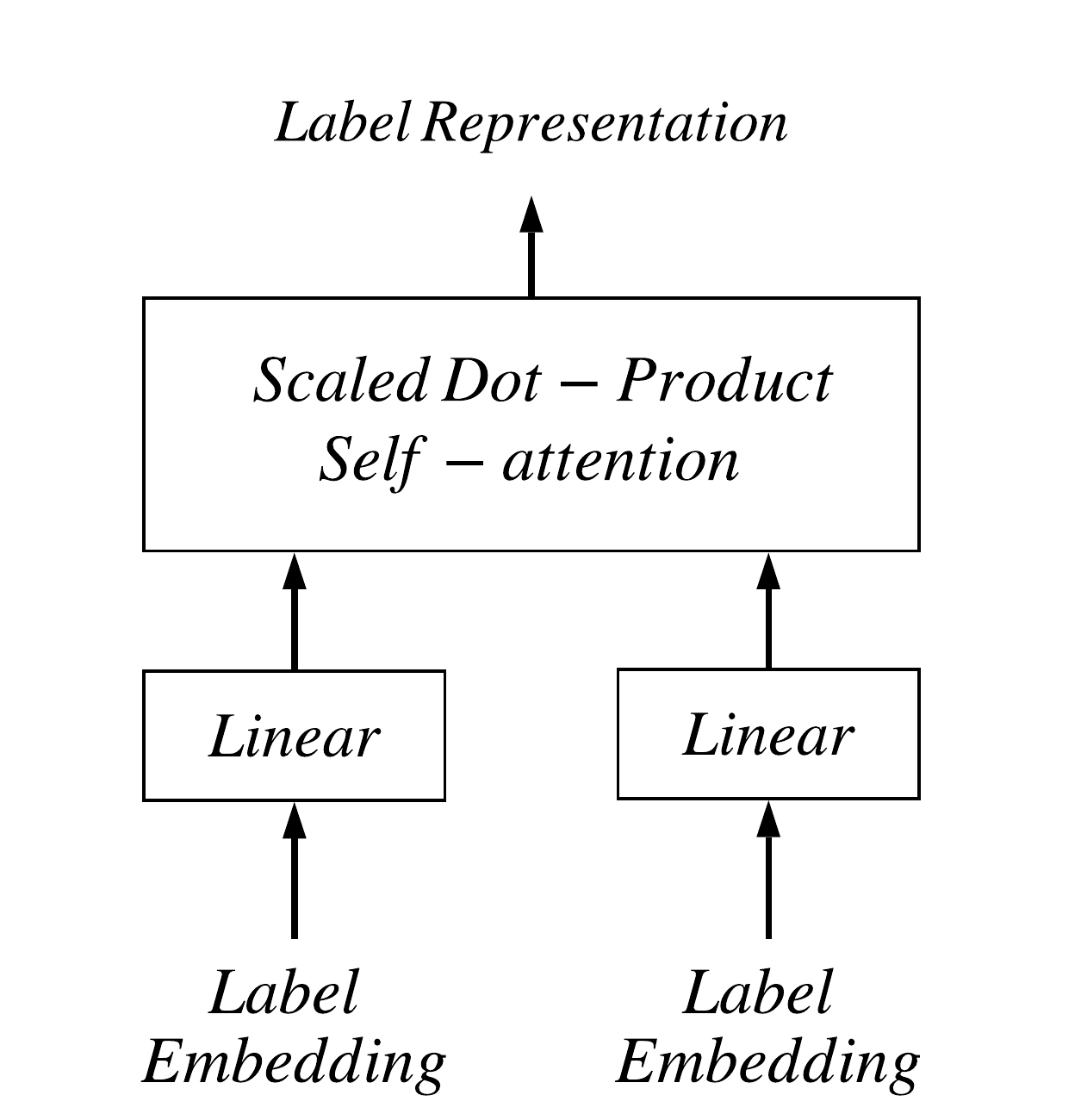}
  \caption{The implementation detail of self-attention.}
  \label{figure3}
\end{figure}

Our implementation is shown in Figure \ref{figure3}. Let $\mathbf{Z}$ represent former connected layers' output \emph{i.e.}, the sum of label embedding and positional encoding. The labels' representations are calculated by the scaled dot-product self-attention:
\begin{equation}
    SelfAttention(\mathbf{Z}, \mathbf{Z}) = SoftMax(\frac{QK^T}{\sqrt{d_z}})\mathbf{Z},
\end{equation}
where $Q, K$ are two different linear maps of $\mathbf{Z}$, and $d_z$ is the dimension of $\mathbf{Z}$.
 
\subsubsection{\bfseries{Add $\&$ Norm layer}} Adding the residual connection and batch normalization \cite{ioffe2015batch}aims to achieve better generalization performance and more stable training process\cite{shen2018disan}. The normalization avoids the values of model changing too much; and the residual link can restrict the gradient disappearance in the self-attention layer.

\subsubsection{\bfseries{Feed Forward layer}} The label encoder contains a two layer fully connected feed-forward network, with a ReLU activation function after the first linear layer:
\begin{equation}
    FFN(x) = Max(0, xW_1 + b_1)W_2 + b_2.
\end{equation}
The inner dimension of the $FNN$ is $d_{ff} =2048$.

\begin{figure}[!tb]
  \centering
  \includegraphics[width=0.49\textwidth]{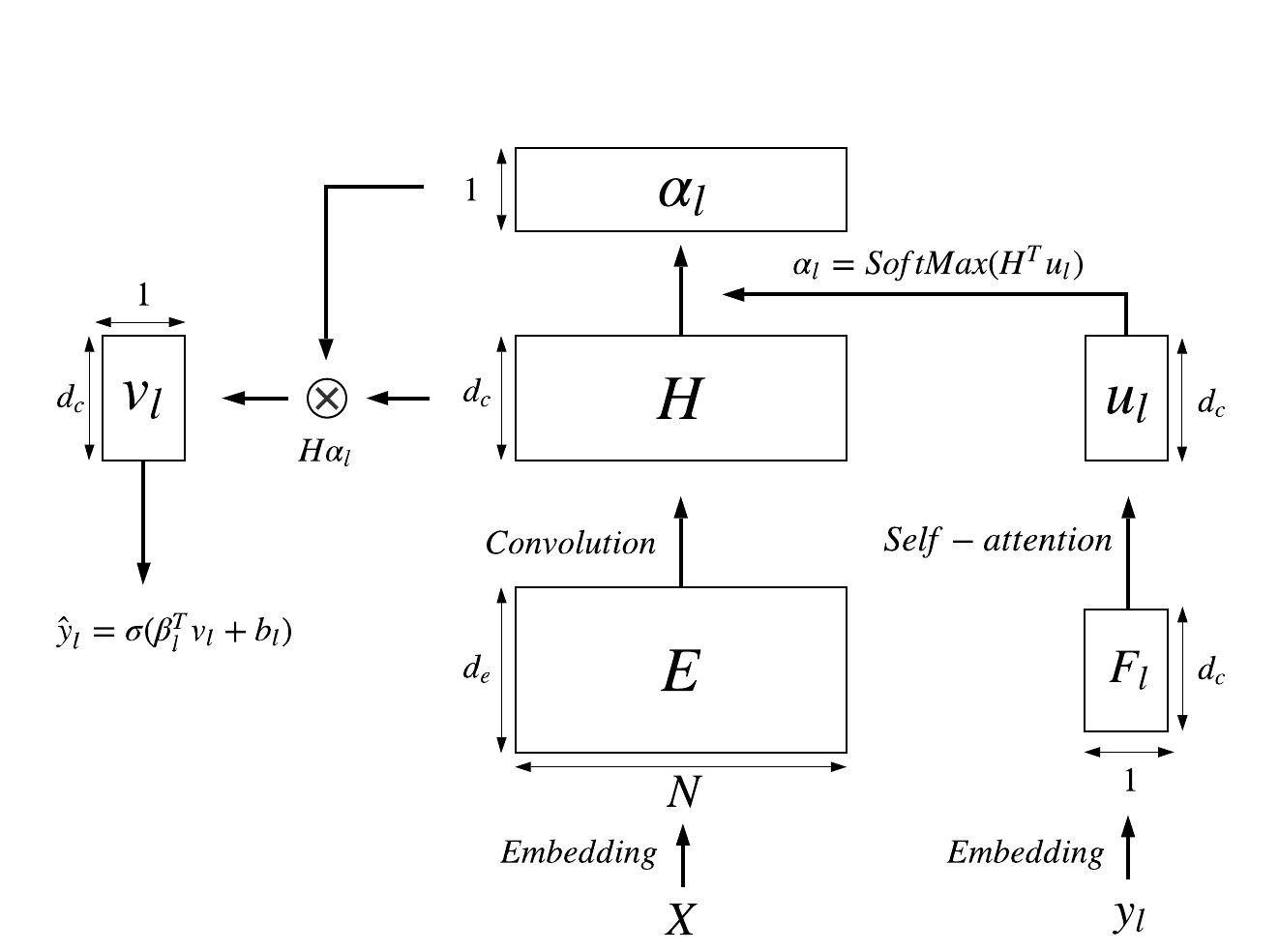}
  \caption{MVAM overall architecture with per-label soft attention alignment.}
  \label{figure4}
\end{figure}

\subsection{Attentional Match}
The overall model with detail implementation is illustrated in Figure \ref{figure4}. We denote the output of the convolutional layer as $\mathbf{H}\in \mathcal{R}^{d_c \times N}$. The input sequence length $N$ is various for different training samples, which is hard to be used by a standard classifier. An usual solution is to reduce the context matrix $\mathbf{H}$ to a representation vector via pooling across the length of input document\cite{kim2014convolutional}, \emph{i.e.} choose the maximum value or the average at each row of $\mathbf{H}$. In very deep CNNs setting, this mechanism is verified to keep rotation invariant and hopefully to get a robust model\cite{lecun2015deep}. Although pooling is effective, it is a too coarse-grained method under our encoder architecture. Considering that the encoder only contains a single convolutional layer, maxpooling or averagepooling can not fully utilize the internal context information of medical documents. A widely used alternative is the attention approach, which focus on a subset of semantic fragments in a document. In addition, the ICD code prediction is an instance of multi-labels classification. We apply a per-label soft attention\cite{mullenbach2018explainable,shen2018disan} to calculate a weight average across $\mathbf{H}$ w.r.t. the given label. This approach selects the k-grams semantic snippets from the text that are most
relevant to each clinical code. A formal expression is descried below.

For any input document $\mathbf{X}$ and label $l$, we first encode them into corresponding representation $\mathbf{H}$ and $u_l\in\mathcal{R}^{d_e}$. Then pass their inner product $\mathbf{H}^Tu_l$ into a softmax operator to calculate relevance distribution over input sequence,
\begin{eqnarray}
  \alpha_l &=& SoftMax(\mathbf{H}^Tu_l),\\
    SoftMax(x) &=& \frac{exp(x)}{\sum_i exp(x_i)}  
\end{eqnarray}
The output attention scores $\alpha_l$ work as weights to calculate the  representation vector $v_l$ of given document $\mathbf{X}$ and label $l$.
\begin{equation}
   v_l = \sum_{n=1}^N \alpha_{l,n}\,h_n,
\end{equation}
where $h_n$ is the $n^{th}$ column of matrix $ \mathbf{H}$.

\subsection{Classifier}
When getting the representation vector $v_l$, we feed it into a single layer classifier with sigmoid activation to predict the probability of given label $l$.
\begin{equation}
    \hat{y_l} = \sigma(\beta_l^Tv_l + b_l),
\end{equation}
where $\beta_l$ is trainable weights, and $b_l$ is a scalar bias.

\subsection{Training}
We employ binary cross entropy as training loss for each ICD code\cite{rubinstein2013cross}, and the training process minimizes the BCE loss via stochastic gradient descendent\cite{bottou2010large}.
\begin{equation}
    L_{BCE}(\mathbf{X},\mathbf{Y}) = -\sum_{l=1}^{\mathcal{L}}{y_l}\log(\hat{y_l})+{(1-y_l)}\log(1-\hat{y_l})
\end{equation}

\section{Evaluation}
This section evaluates the effectiveness of MVAM on the public dataset and compare its performance against several competitive baselines and the state-of-art model. 
\subsection{Dataset}
We perform the study on an open source dataset MIMIC-III \cite{johnson2016mimic}. MIMIC-III contains data associated with 53,423 distinct hospital admissions for adult patients (aged 16 years or above) admitted to critical care units between 2001 and 2012. The data covers 38,597 distinct adult patients and 49,785 hospital admissions. We Follow the experiment setting in \cite{mullenbach2018explainable} and focus on discharge summaries, which condense information about a stay into a single document. Then summary and addenda are concatenated  into one document.

Each visit record is marked by a coder with a set of ICD-9 codes that describe the diagnosis and procedures during the patient's hospital stay. There are 8,921 unique ICD-9 codes in our dataset. the original data are split by patient ID to avoid the same patient visits appear in both the training and test sets. To guarantee the fairness among comparison methods, use the dataset in two settings, the full-label setting and the 50 most frequency labels setting.
In the full-label setting (MIMIC-III, full labels), we use a set of 47,724 discharge summaries for training, 1,632 summaries for validation and 3,372 for testing.

In the most frequent setting (MIMIC-III 50), we follow \cite{shickel2017deep} and exclude the non-frequent samples in the above datasets. The statistic of these two setting is shown in table \ref{table1}.
\begin{table}
  \centering
  \caption{Statistics for MIMIC-III discharge summary training sets.}
    \begin{tabular}{lcc}
    \toprule
          & \textbf{MIMIC-III full} & \textbf{MIMIC-III 50}  \\
    \midrule
    training samples & 47,724 & 8,067\\
    Vocabulary size & 51,917 & 51,917  \\
    Mean \# tokens & 1,485 & 1,530 \\
    Mean \# labels  & 15.9  & 5.7   \\
    Total \# labels & 8,921 & 50 \\
    \bottomrule
    \end{tabular}%
  \label{table1}%
\end{table}%
\subsection{Comparison Methods}
We set up three baselines for two dataset settings:
\begin{itemize}
    \item {single-layer convolutional
neural network\cite{kim2014convolutional}}.
    \item {bag-of-words logistic regression model}.
    \item {bidirectional gated recurrent unit (Bi-
GRU)\cite{jagannatha2016bidirectional}}.
\end{itemize}
We also compare with several typical prior methods, C-MemNN \cite{prakash2017condensed}, Shi \emph{et al.}\cite{shickel2017deep} on MIMIC-III 50. Especially, we compare MVAM against CAML and DR-CAML\cite{mullenbach2018explainable}, which achieve state-of-art results on the MIMIC-III clinical codes prediction.

In all experiments with words' distributed representation, we use the same pre-trained 100-dimensional word2vec embedding layer. For fairness, we set our label embedding layer at 100 dimension, and initialize it randomly. The experiments are implemented as Pytorch projects.

\subsection{System Setup}
As point out in section 1, our framework highlights its efficiency on the easier training setup. MVAM only includes 4 hyper parameters, they are the CNN output channels $d_c$, convolutional kernel width $k$, the dropout rate $p$ for word embedding layers and the initial learning rate $\eta$ for adam\cite{kingma2014adam} optimizer. It dose not induce any hyper parameters to constrain the model as described in \cite{mullenbach2018explainable}, or any trade off rate to force the label encoder to learn an isomorphic representation as the input.

Moreover, MVAM is insensitive to values of hyper parameters, except for the dropout rate $p$ in the input encoder, usually we allocate a quite large one $p=0.6$ for MIMIC-III full labels setting and $p=0.8$ for MIMIC-III 50 labels setting. In our experiments, we fix all other hyper parameters at $d_c = 200$, $k = 10$, and $\eta = 0.0002$.  Thus, even though we train a little longer than the comparison method CAML on a single running, we could still save much more effort on tuning the hyper parameters. 

We apply the early stop strategy to avoid model over fitting. On the MIMIC-III full data, we choose precision $@$15 on validation set as early stop criteria. Model training will be terminated when the precision $@$15 does not increase for 10 epochs. The model with the highest value of  precision$@$15 is evaluated on the test data. On the MIMIC-III 50 data, we use the same policy except for using precision$@$5 as early stop criteria.

The experiments are conducted using Pytorch, on a CentOS 7 system equipped with 4 NVIDIA Tesla P40 graphic cards. 
\subsection{Evaluation Metrics}
To fairly and fully compare with prior works, we report all metrics listed in\cite{mullenbach2018explainable}, including the micro-averaged and macro-averaged F1 score, area under the ROC curve (AUC) and precision at n('P$@$n'), which is the fraction of the n highest scored labels that are present in the ground truth. Mullenbach~\emph{et al.}~\cite{mullenbach2018explainable} claim that it is motivated by the potential use case as a decision support application, in which a user is presented with a fixed number of predicted codes to review.

Micro-averaged
values are calculated by treating each (text, code) pair as a separate prediction. Macro-averaged values are calculated by averaging metrics computed per-label. For $F1$ score, the metrics are distinguished as follows:
\begin{eqnarray}
    Micro-F1 &=& \frac{\sum_{l=1}^{|\mathcal{L}|}2TP_l}{\sum_{l=1}^{|\mathcal{L}|}2TP_l + FP_l + FN_l }\\
    Macro-F1 &=& \frac{1}{|\mathcal{L}|}\sum_{l=1}^{|\mathcal{L}|}{\frac{2TP_l}{2TP_l+FP_l+FN_l}},
\end{eqnarray}
where TP denotes true positive, FP denotes false positive and FN denotes true negative. All other metrics are calculated analogously.

\subsection{Quantitative Results}
\begin{table*}[htbp]
  \centering
  \caption{Quantity results on MIMIC-III, full labels.}
    \begin{tabular}{l|cc|cc|cc}
    \toprule
          & \multicolumn{2}{c|}{AUC} & \multicolumn{2}{c|}{F1}  & \multicolumn{2}{c}{P@n} \\
    Model & Macro & Micro & Macro & Micro    & 8     & 15 \\
    \midrule
    Logistic Regression & 0.561 & 0.937 & 0.011 & 0.272 & 0.542 & 0.411 \\
    CNN   & 0.806 & 0.969 & 0.042 & 0.419 & 0.581 & 0.443 \\
    Bi-GRU & 0.822 & 0.971 & 0.038 & 0.417 & 0.585 & 0.445 \\
    CAML  & 0.895 & \textbf{0.986} & 0.088 & 0.539  & \textbf{0.709} & 0.561 \\
    DR-CAML & 0.897 & 0.985 & 0.086 & 0.529 & 0.690  & 0.548 \\
    MVAM  & \textbf{0.911} & \textbf{0.986} & \textbf{0.094} & \textbf{0.549} & {0.708} &\textbf{0.565}  \\
    \bottomrule
    \end{tabular}%
  \label{table2}%
\end{table*}%

To evaluate the scalability and effectiveness of MVAM, the first quantitative experiment predicts all 8921 labels on the MIMIC-III full dataset. Our results are reported in table \ref{table2}, as well as results published in \cite{mullenbach2018explainable}.
The MAVM model achieves the best results on all metrics with the only exception of P$@$8 by 0.001. Especially for the macro metrics, MAVM outperforms CAML ,as well as its enhanced version DR-CAML, with Statistical significance. As we know, the macro-averaged metrics emphasis on rare label prediction\cite{van2013macro}. 

This result reveals the excellent performance of MVAM on rare seen labels prediction, even compared with the specifically designed architecture DR-CAML \cite{mullenbach2018explainable}, which is trained with auxiliary text description of ICD codes. We credit this improvement to self-attention and positional encoding. Because of the limited samples, the embedding of the rare labels does not update as frequently as others. The self attention in label encoder relates the rare labels to the frequent ones, which can be seen as anchors to separate the rare ones from each other. 

We train MVAM without positional encoding in full labels setting, the performance drop dramatically. We would like to explain this situation from the perspective of long distance dependency. Self attention module needs to compute one versus all inner product, then execute weight average along all labels. As the sequence length increasing, this value will degenerate to mean over all label embedding, and the gradient will disappear in the label encoder, as a result, the model will suffer the less fit condition. Under this situation, the sequence is too long to be handled alone by self-attention. The encoded positional information can help the self attention module distinguish more different labels.

To compare with published work, we then evaluate on most frequent ICD codes, the results are reported in table \ref{table3}. In the setting of MIMIC-III, 50, we can observe significant improvement over all comparison methods in all reported metrics. In particular, MVAM makes great progress in Micro AUC, Micro F1 cores, and P$@$5. Contrary to Macro average, the Micro metric emphasis on frequent labels' prediction\cite{chaudhari2019attentive}.

We omit the positional encoding unit in this experiment, since there is only 50 labels to predict. Therefore, we believe that label inner alignment does encode spacial structure information into the final label representation, and this information helps the input decoder to find a better way to distill meaningful snippets from its text representation.

\section{Related Work}
In this section, we will introduce some representative related works, which give us hint to implement the MVAM. Basically we introduce their novelty, strength and weakness, commonality and difference with us.

\subsection{Attention Models}
Attention is somehow new created term in machine learning community. It is considered as a branch of alignment\cite{deng2018latent} by some researchers. Even in the work\cite{bahdanau2014neural} which first applies attention to NLP field, the authors still use the word alignment. The attention block is usually implemented as a neural network. 

The intuition behind attention can be best explained using human biological systems\cite{chaudhari2019attentive}. For instance, we can understand others by paying attention to words related to the current topic. 

Generally speaking, the attention block is responsible for automatically learning weights of underlying representation fragments by their semantic importance. The mainstream is using the attention block as a connection component in an encoder-decoder framework. Yang \emph{et al.}\cite{yang2016hierarchical} propose a two layers GRU based attention network for document classification. The attention block is used in both word embedding layer and sentence embedding layer. In \cite{yin2017attentive}, Yin \emph{et al.} applies a similar attention of RNN into a convolutional neural network. 

Our model inherits the attention concept from aforementioned methods, but replace the underlying attention block by a inner product, which simplifies implementation and training processes.

\subsection{Label Embedding}
Label embedding , as the name suggests, generates a distributed representation for given label, rather than $\{0,1\}$. It utilizes the label correlation captured in the embedding space to improve the prediction performance. In NLP, labels embedding for text classification has been widely studied. Our framework breaks the limitation of parallel computing by abandoning the classic RNN encoder architecture.

Wang \emph{et al.}\cite{wang2018joint}design an attention model, joint embedding of words and labels to make use of label information for text classification. The pipe line of this method is similar to us, but it tries to map the labels and documents into the same embedding space, which restricts the exploration ability of their model. 

Another work \cite{shi2017towards} proposed a hierarchical embedding RNN model combined with attention to predict ICD diagnostic codes. This work gives some hint for us to detect the direct correlation in the label space. The recurrent structure limits their parallel ability, and they can not expand to large label space.

The CAML\cite{mullenbach2018explainable} is the most relative model to ours. In contrast to \cite{wang2018joint}, they turn to align the representation space, which is more flexible than the alignment on embedding space.

We share the same input encoder and attention module structure with CAML. However, we observe a dilemma in their expression. DR-CAML encodes the text description of ICD code as the label embedding for a given label, and trains word embedding and label embedding simultaneously. They perform worse than CAML whose labels are randomly initialized. We solve this problem by adding self attention label encoder to discover direct interactions among labels. Moreover, DR-CAML prefers homogeneous encoder and decoder framework, while in our model, heterogeneous input and label encoders are incorporated. 

\subsection{Self Attention \& Transformer}
Self attention based Transformer \cite{vaswani2017attention} makes a great contribution to the progress of NLP tasks. BERT\cite{devlin2018bert} and GPT\cite{radford2018improving} are all built upon the transformer structure. Transformer fully uses self attention and feed forward neural networks to learn text representation. Since it breaks the limitation of recurrent dependency, the training can be employed in a much larger scale. Also the performance of downstream tasks has proved its effectiveness. Our label encoder is similar to the architecture of Transformer. we change the structure a little bit. we remove the multi-head module and introduce a much larger positional encoding (8921) which is at most 512 in the original setting.

\section{Conclusion}
We propose the MVAM, an elegant solution for automatic ICD codes prediction. It employs a label inner alignment encoder to discover semantic structures among labels. Practical engineering designs are induced, which make our framework efficient and effective. MVAM yields significant improvements over state-of-art works on public ICD-9 code prediction tasks. Although the MVAM is design for clinical setting, it is extensible without any modification to other multi-label document annotating tasks.
\begin{table}
  \centering
  \caption{Quantity results on MIMIC-III, 50 labels.}
    \begin{tabular}{l|cc|cc|c}
    \toprule
          & \multicolumn{2}{c|}{AUC} & \multicolumn{2}{c|}{F1} &  \\
    Model & Macro & Micro & Macro & Micro & \multicolumn{1}{c}{P@5} \\
    \midrule
    C-MemNN \cite{prakash2017condensed} & 0.833 &   ---  &   ---    &    ---   & 0.420 \\
    Shi et al.\cite{shickel2017deep} &  ---     & 0.900   &   ---    & 0.532 & --- \\
    Logistic Regression & 0.829 & 0.864 & 0.477 & 0.533 & 0.546 \\
    CNN   & 0.876 & 0.907 & 0.576 & 0.625 & 0.620 \\
    Bi-GRU & 0.828 & 0.868 & 0.484 & 0.549 & 0.591 \\
    CAML  & 0.875 & 0.909 & 0.532 & 0.614 & 0.609 \\
    DR-CAML & 0.884 & 0.916 & 0.576 & 0.633 & 0.618 \\
    MVAM  & \textbf{0.896} & \textbf{0.928} & \textbf{0.580} & \textbf{0.661} & \textbf{0.634}  \\
    \bottomrule
    \end{tabular}%
  \label{table3}%
\end{table}%

\section{Conclusion}
The conclusion goes here.


%

\ifCLASSOPTIONcaptionsoff
  \newpage
\fi



\bibliographystyle{IEEEtran}
\bibliography{IEEEabrv,IEEEexample.bib}
%




%

\begin{IEEEbiography}{Guodong Liu}
A ph.D student in the Univerisity of Pittsburgh.
\end{IEEEbiography}







\end{document}